\documentclass{article}

\usepackage{PRIMEarxiv}

\usepackage[utf8]{inputenc} 
\usepackage[T1]{fontenc}    
\usepackage{url}            
\usepackage{booktabs}       
\usepackage{amsfonts}       
\usepackage{nicefrac}       
\usepackage{microtype}      
\usepackage{lipsum}
\usepackage{fancyhdr}       
\usepackage{graphicx}       
\graphicspath{{media/}}     

\usepackage{amsmath}
\usepackage{amssymb}

\usepackage{ragged2e} 
\usepackage{booktabs,makecell, multirow, tabularx}

\usepackage{pifont}
\newcommand{\cmark}{\ding{51}} 
\newcommand{\xmark}{\ding{55}} 

\usepackage{xcolor}
\definecolor{LightGray}{gray}{0.85}
\definecolor{LightBlue}{rgb}{0.97,0.985,1.0}
\definecolor{LightOrange}{rgb}{1.0,0.97,0.94}
\definecolor{LightGreen}{rgb}{0.98,1.0,0.98}
\definecolor{LightPurple}{rgb}{0.96,1.0,1.0}
\definecolor{dark-gray}{gray}{0.30}
\newcommand{\pub}[1]{{\color{dark-gray}{\scriptsize{[{#1}]}}}}

\usepackage[most]{tcolorbox}

\usepackage{threeparttable}

\usepackage[round,authoryear]{natbib}
\usepackage{hyperref}       

\title{
MerNav: A Highly Generalizable Memory–Execute–Review Framework for Zero-Shot Object Goal Navigation
}

\author{
\hspace{0.75em} Dekang Qi\textsuperscript{\rm 1},
Shuang Zeng\textsuperscript{\rm 2, 1}\thanks{is an intern at Amap, Alibaba Group},\hspace{0.75em}
Xinyuan Chang\textsuperscript{\rm 1}, \hspace{0.75em} Feng Xiong\textsuperscript{\rm 1}, \hspace{0.75em}  \\
\textbf{
Shichao Xie\textsuperscript{\rm 1}, \hspace{0.75em}
Xiaolong Wu\textsuperscript{\rm 1}, \hspace{0.75em}
Mu Xu\textsuperscript{\rm 1}
}\\
\textsuperscript{\rm 1}Amap, Alibaba Group \hspace{0.5em} \textsuperscript{\rm 2}Xi’an Jiaotong University \\
\tt\small
\tt\small \{qidekang.qdk, changxinyuan.cxy, huanlu.wxl, xumu.xm\}@alibaba-inc.com, \\
\tt\small zengshuang@stu.xjtu.edu.cn \{xf250971, tenan.xsc\}@autonavi.com
}

\begin{document}

\bibpunct{(}{)}{;}{a}{,}{,}

\renewcommand{\harvardand}{\&}
\let\cite\citep

\maketitle

\begin{abstract}
Visual Language Navigation (VLN) is one of the fundamental capabilities for embodied intelligence and a critical challenge that urgently needs to be addressed.
However, existing methods are still unsatisfactory in terms of both success rate (SR) and generalization: Supervised Fine-Tuning (SFT) approaches typically achieve higher SR, while Training-Free (TF) approaches often generalize better, but it is difficult to obtain both simultaneously. 
To this end, we propose a Memory-Execute-Review framework.
It consists of three parts: a hierarchical memory module for providing information support, an execute module for routine decision-making and actions, and a review module for handling abnormal situations and correcting behavior.
We validated the effectiveness of this framework on the Object Goal Navigation task. Across 4 datasets, our average SR achieved absolute improvements of 7\% and 5\% compared to all baseline methods under TF and Zero-Shot (ZS) settings, respectively.
On the most commonly used HM3D\_v0.1 and the more challenging open vocabulary dataset HM3D\_OVON, the SR improved by 8\% and 6\%, under ZS settings. Furthermore, on the MP3D and HM3D\_OVON datasets, our method not only outperformed all TF methods but also surpassed all SFT methods, achieving comprehensive leadership in both SR (5\% and 2\%) and generalization. Additionally, we deployed the MerNav model on the humanoid robot and conducted experiments in the real world. The project address is: https://qidekang.github.io/MerNav.github.io/
\end{abstract}


\section{Introduction}
\begin{figure}[t]
  \centering
  \includegraphics[width=0.5\linewidth]{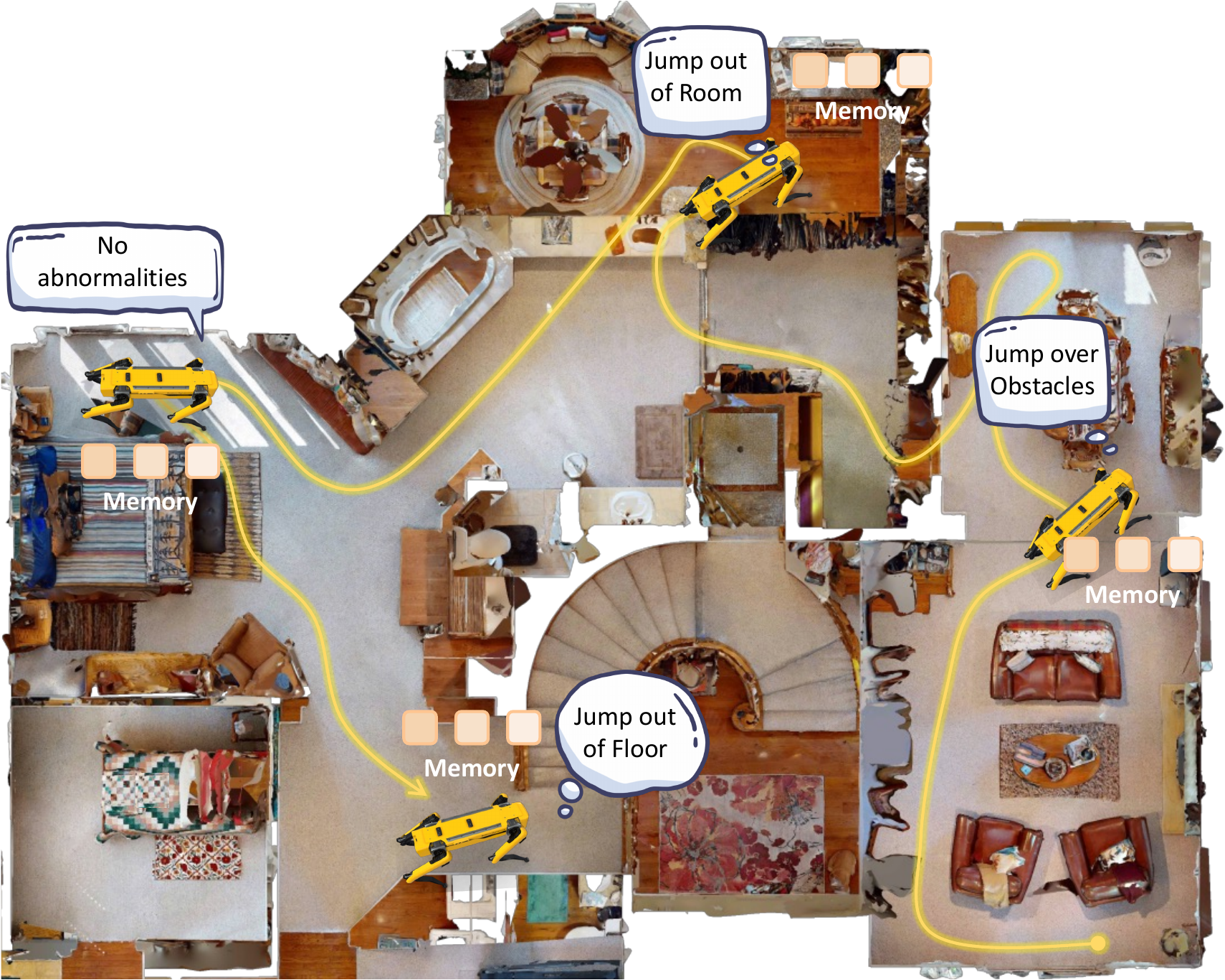}
  \caption{Our Memory-Execute-Review VLN framework, supported by a hierarchical memory structure, can handle regular and exceptional situations using the execute and review modules.}
  \label{fig:motivation}
\end{figure}
Embodied intelligence has enormous application prospects, and it is predicted to reach a market size exceeding 5 trillion dollar in the future \cite{MorganStanley2025Humanoids5T}, making it a shared focal point for both academia and industry.
Among the key capabilities required by embodied intelligence, VLN constitutes one of the most fundamental yet most urgent bottlenecks.
Without reliable navigation, the practical viability of embodied intelligence would be fundamentally constrained; in an extreme sense, embodied intelligence risks becoming a “castle in the air.”
Nevertheless, existing VLN models remain unsatisfactory in terms of SR and generalization.
Existing VLN approaches can be broadly categorized into two lines.
SFT VLN: This line adapts a foundation model using task-specific training data and typically yields slightly higher SR. However, it often induces capability degradation of the foundation model and leads to overfitting to a particular data distribution, thereby weakening generalization.
TF / Agentic VLN: This line largely preserves the general competence of the foundation model by introducing multi-agent collaborative pipeline and integrating domain knowledge and task characteristics. It typically offers stronger generalization, but its SR is often slightly lower than that of SFT approaches.

Our intuition is that the weak generalization and high data demand inherent to the SFT VLN paradigm are difficult to eliminate in the near term. In contrast, TF VLN offers advantages including stronger generalization, lower data requirements, enhanced tool use and error-correction capability, and improved interpretability, suggesting greater long-term potential. Furthermore, a hybrid paradigm based on foundation models, using the TF method as the core scaffold and the SFT method as a supplement, may emerge as an even more compelling direction.
Despite these advantages, current TF/agentic VLN methods remain at an early stage in modeling human cognition. They typically reproduce only limited and simplified cognitive modules, which constrains overall performance. The limitations mainly arise from incomplete modeling of memory systems and oversimplification of thought processes.

Inspired by the collaborative mechanisms of the human brain and human groups, we propose a highly generalizable \textbf{\underline{M}}emory-\textbf{\underline{E}}xecute-\textbf{\underline{R}}eview framework for \textbf{\underline{Nav}}igation (\textbf{MerNav}, where mer means sea in French and also serves as an English prefix related to the sea). As illustrated in Fig. \ref{fig:motivation}, MER is supported by a hierarchical memory structure, can handle regular and exceptional situations specifically using the execute and review modules.
We further validate MER on the Object Goal Navigation task, which more aligned with real world human–machine interaction scenarios. To the best of our knowledge, this is the first model that achieves comprehensive leadership in success rate, generalizability, and interpretability across multiple datasets and various (TF and SFT) methods.
Our main contributions are:
\begin{itemize}
\item 
Propose the Memory–Execute–Review VLN framework, improving the model’s memory capability and its ability to handle routine and anomalous situations.
\item 
We construct a hierarchical memory system comprising short-term, long-term, and commonsense memory to provide rich informational support for each module. We further design a procedural pipeline: observation analysis, path planning, action selection, stop decision, to enhance robustness under routine conditions. In addition, we introduce two-step and multi-step review mechanisms based on process information to strengthen error correction in anomalous situations.
\item 
Across 4 datasets, the SR improved by an average of 7\%, and 5\% compared to all methods under the TF and Zero-Shot setting, respectively. On the most commonly used HM3D\_v0.1 and the newer, more challenging open vocabulary dataset HM3D\_OVON, the SR improved by 8\% and 6\% in Zero-Shot Setting, respectively.
Furthermore, in both MP3D and HM3D\_OVON, it outperformed not only all TF but also all SFT methods, achieving overall leading performance in both success rate (5\% and 2\%) and generalization ability.
Adopting a more advanced foundation model can further boost SR and SPL by 3\% each, bringing the Success Rate to a "Acceptable" level ($ > $70\%).
\end{itemize}

\section{Methodology}
\subsection{Task Definition}
Object Goal Navigation is defined as follows:
Initially, the embodied agent is placed at a designated start location $P_0$ and receives an instruction specifying a goal $G \in \mathcal{N}$, such as ``bed'', ``chair.'' At time step $t$, the agent's observation $ O_t $ consists of an RGB-D image $I_t$ and its position and pose $P_t$ in the environment $\mathcal{E}$. Based on the current input, the model outputs the expected action $A^e_t$. After the action is executed in the environment, the agent obtains the observation at $t+1$, and this process iterates until the agent chooses to stop. If the distance between the agent's stopping location and the goal is smaller than a threshold $T_{g}$, the task is considered successful; otherwise, it is considered a failure.

\subsection{Overview}
\begin{figure*}[t]
  \centering
  \includegraphics[width=0.95\linewidth]{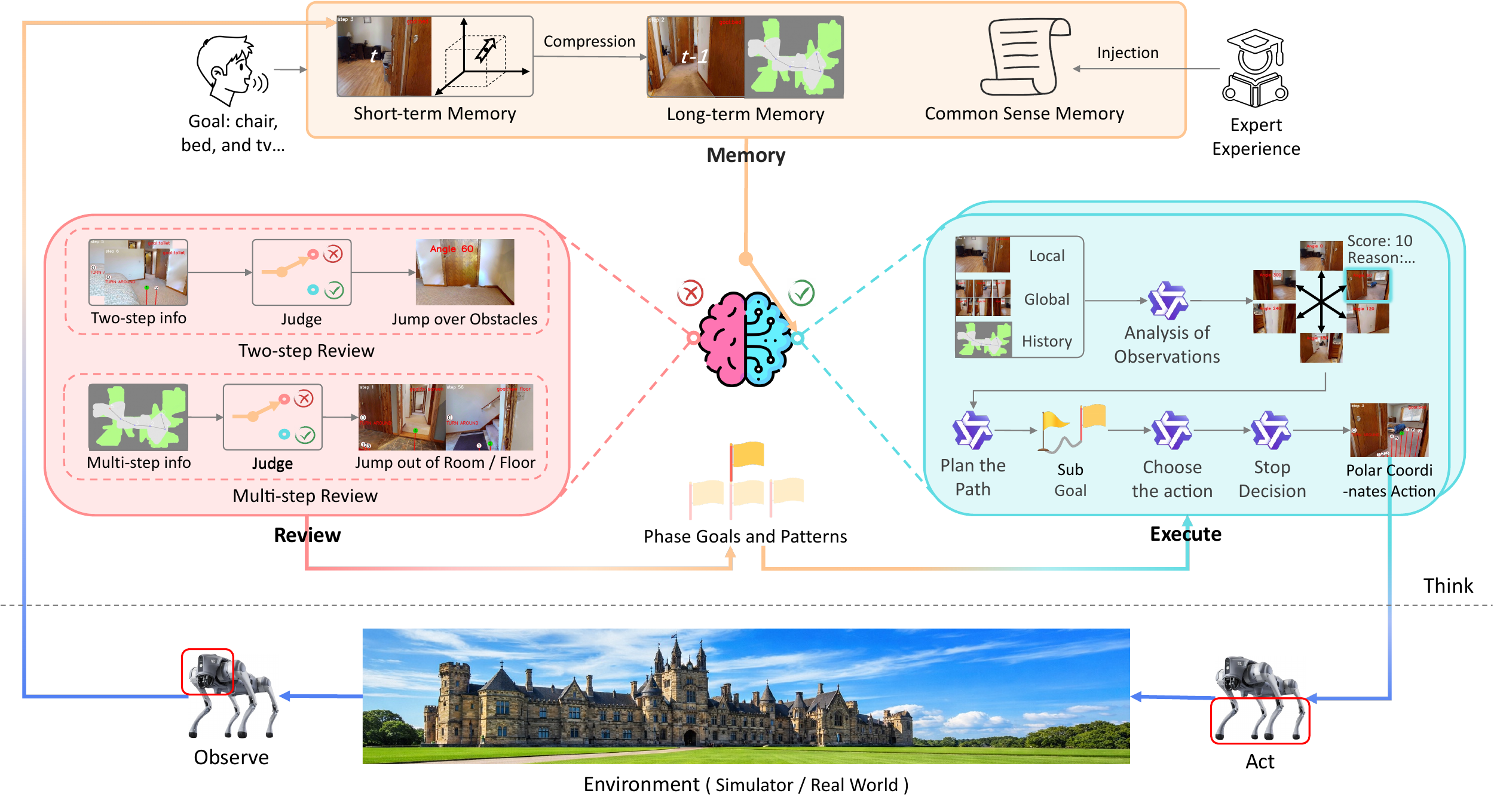}
  \caption{Overview of Memory-Execute-Review.}
  \label{fig:overview}
\end{figure*}
The proposed Memory-Execute-Review (MER) VLN framework is shown in Figure \ref{fig:overview}. The agent’s sensors acquire the observation $ O_t $ of the current state $ S_t $. Conditioned on the user-specified goal instruction $ G $ and $ O_t $, the model through thinking and generates the expected action $A^e_t$. Then the agent’s body instantiates this action as the real action $A^r_t$ in the environment (simulator or real world), which updates the current state to the next state $ S_{t+1} $. This process forms a closed-loop \textbf{Observe–Think–Act} iteration in the environment, as defined in Equation \ref{con:OTA}.
\begin{equation}
\begin{aligned}
Observe:\quad & O_t = f_{Observe}(S_t),\\
Think:\quad & A^{e}_{t} = f_{Think}(G, O_t),\\
Act:\quad & A^{r}_{t} = f_{Act}(A^{e}_{t}),\\
\mathcal{E}:\quad & S_{t+1} = f_{\mathcal{E}}(S_t, A^{r}_{t}).
\end{aligned}
\label{con:OTA}
\end{equation}
The thinking process consists of \textbf{Memory-Execute-Review}. The Memory module provides priors and informational support for decision-making; under nominal conditions, task completion is handled by the Execute module. Meanwhile, the Review module continuously monitors the execution process from an independent perspective and, upon detecting anomalies or deviations, triggers corresponding corrective modes to rectify behavior, as shown in Equation \ref{con:MER}.
\begin{equation}
\begin{aligned}
M_t &= f_{Memory}(G, O_t),\\
G_p &= 
\begin{cases}
f_{Review}(G, M_t), & if abnormal,\\
\varnothing, & otherwise,
\end{cases}\\
A^e_t &= f_{Execute}(G, G_p, M_t).
\end{aligned}
\label{con:MER}
\end{equation}
This functional decomposition bears resemblance to the regional collaboration in the human brain.
On the one hand, some brain regions are responsible for memory storage and retrieval, such as the hippocampus and medial temporal lobe system \cite{tulving1998episodic}, and the dorsolateral prefrontal cortex \cite{barbey2013dorsolateral}.
On the other hand, some regions are better suited for stable processing and control under routine situations, including the default mode network \cite{raichle2015brain}, the frontoparietal network \cite{marek2018frontoparietal}, basal ganglia-based habitual responses \cite{yin2006role}, and automatic information processing in the human brain \cite{schneider1977controlled}. 
In contrast, other regions are more strongly implicated in anomaly detection and conflict processing, such as the anterior cingulate cortex \cite{swick2002dissociation} and the salience network \cite{ham2013cognitive}.
As the ancient philosopher Zengzi said, "I examine myself three times a day," humans often execute and review in parallel when completing tasks, continuously reviewing to correct behavior and improve reliability.
A similar division of labor is also widespread in human collaborative systems. for example, some members write code while others conduct code review; likewise, in the judicial system, the court is responsible for trials, while the prosecution assumes a supervisory role. Such “execute–review” collaboration can improve a system’s safety and robustness while maintaining efficiency.

\subsection{Memory}
In neuroscience research, memory is commonly categorized using several classic taxonomies. For example, memory can be divided into short-term memory and long-term memory: short-term memory further includes sensory memory and working memory; long-term memory can be divided into explicit memory and implicit memory, where explicit memory further contains episodic memory and semantic memory, while implicit memory mainly corresponds to procedural memory \cite{wu2025human}. In addition, some studies categorize memory by time scale into short-term memory, long-term memory, and long-lasting memory \cite{lechner1999100}\cite{roselli2021making}. 
Some work has also examined common sense memory \cite{akhtar2018common}.
These categorizations emphasize that different types of memory differ in terms of their information sources, storage mechanisms, accessibility, and functional roles.
Inspired by this, we construct a hierarchical memory system composed of short-term memory, long-term memory, and common-sense memory to simulate the structural characteristics of human memory. We regard “Thinking” as the process of operating on memory; therefore, memory plays a crucial role. Compared with approaches that rely only on current inputs or that introduce only simple long-term information, we enrich and structure the contents of long-term memory, and introduce common-sense memory via expert knowledge injection, making behavioral control more intuitive and interpretable.

\textbf{Short-term Memory.} It represents instantaneous information directly obtained from the external world at the current moment $ O_t $, including RGB-D images $ I_t $ and the embodied device’s position and pose information $ P_t $.

\textbf{Long-term Memory.} It is task-specific memory that is cleared once the task ends. Constrained by storage capacity and the efficiency of retrieval and processing, long-term memory cannot grow without bound, and we divide it into compressed and uncompressed information.

\textit{Compressed Memory.}
It is used to retain key structured information with low storage overhead, including the exploration-area record map and the exploration-value record map.
The exploration-area record map is a bird’s-eye-view (BEV) map constructed based on depth maps and poses information (without relying on an accurate scene graph or occupancy map), and it can mark areas in different colors according to the exploration status.
We assume that nearby areas within the agent’s field of view can be observed more reliably. Therefore, regions within an observation distance threshold $T_o$ are considered explored and are marked in gray, whereas regions beyond $T_o$ may be insufficiently observed and are treated as unexplored and marked in green. In addition, black indicates unreachable areas, and dots denote previously visited positions.
The exploration-value record map is also represented in BEV. At each step, the model scores images observed in multiple directions. Because these scores have persistent value and should not be discarded after one-time use, we compress and fuse the scoring information in temporal order to form a distribution of exploration values over directions on the BEV map.

\textit{Uncompressed Memory.}
It is used to preserve task-critical information that we wish to record losslessly. To control storage size, we adopt a time-window mechanism to limit its capacity. For example, the observation image from $ t-1 $ step is often particularly critical for current decision-making, so we store it using a sliding window of length $1$. The sliding-window length is denoted as $L_w$.

\textbf{Common-sense memory.} It is cross-task and persistent, used to accumulate stable experience and rules, and to support experts injecting knowledge into the system (human-in-the-loop). Common-sense memory includes self common sense (e.g., the robot’s own width), goal common sense (e.g., the difference between chair and sofa), and environmental common sense (e.g., cannot pass through a closed door).

In terms of usage, short-term memory and long-term memory are both task-related information and serve as inputs to the model; common-sense memory is reused across tasks and is better injected into the model prompt as rules/priors.
In terms of the correspondence between memory types, common-sense memory is primarily manifested as semantic memory; long-term memory and short-term memory are mainly episodic memory, and together they constitute explicit memory. Meanwhile, the underlying foundation model can be viewed as containing implicit memory learned from large-scale data and stored in parameters. 
\subsection{Execute}
We regard “Thinking” as the process of operating on memory. This process can be divided into an execute module responsible for decision-making and action, and a review module responsible for monitoring and error correction.

The execute module is designed for most normal scenarios and implements a coarse-to-fine decision-making pipeline through multi-modal information fusion. Overall, it consists of four stages: observation analysis, path planning, action selection, and stop decision.

\textbf{Observation Analysis.}
Human vision has an “automatic zoom” capability. Inspired by this, we design an observation analysis stage that integrates information from the local view, global view, and historical view to build a more complete understanding of the environment.

Specifically, Local-view information is the observation image in the current facing direction, used to accurately identify fine-grained details such as target objects and obstacle boundaries, thereby strengthening the model’s attention to that direction. 
Global-view information is formed by stitching images from six directions $\{0^\circ, 60^\circ, 120^\circ, 180^\circ, 240^\circ, 300^\circ\} $, and annotating the corresponding directions on the image. This gives the model more complete spatial-structure information and helps avoid misjudgments caused by limited local view. For example, the local view may only show part of a sofa, which may be mistaken as a bed; a global view showing the full sofa shape enables more stable recognition as a sofa.
Historical-view information is the exploration-
area record map that marks explored and unexplored regions, used to suppress redundant exploration and suggest more promising directions.
We denote \(I_{local}, I_{global},\) and \(I_{history}\) collectively as \(I_{integ}\). As shown in Eq. \ref{con:view}, the observation-analysis agent scores each direction based on $ I_{integ} $ and provides the reasons for its scores. Here, reason $R_t $ explicitly presents the analysis process, improving decision transparency and interpretability.
\begin{equation}
Score^{cur}_t, R_t = f_{AnalysisAgent}(G,G^t_p,I_{integ})
\label{con:view}
\end{equation}
Specifically, the $ f_{AnalysisAgent} $ is implemented based on a VLM foundation model, and its prompt uses a modular structure design, including sections such as objective description, input specification, scoring criteria, commonsense constraints, output guidelines, and output examples, thereby ensuring a more reasonable and controllable analysis and scoring process.
Subsequently, we use the Exponential Moving Average to fuse the current score with the historical score stored in the  exploration-value record map, obtaining a fused score $ Score_t $, giving higher weight to recent information while exponentially decaying the weight of older information.
Among the six directions $ \mathcal{D} $, we select the direction with the highest score as the coarse-filter result:
\begin{equation}
D^t_{sel}= \arg\max_{i\in \mathcal{D}}\ {Score}^i_t
\label{con:Direction}
\end{equation}
The image corresponding to this direction is $ I^t_{sel} $.

\textbf{Path Planning.}
The path-planning module decomposes the high-level goal into a sequence of subgoals, serving as an intermediate layer between coarse direction filtering and fine-grained action selection.
In Object Goal Navigation, only the final goal instruction is provided, while reaching the target may require several to dozens of steps. It is therefore necessary to break the final goal into subgoals that provide executable and verifiable intermediate guidance for subsequent decision-making.
As shown in Equation \ref{con:SubGoal}, we feed such as the currently selected image and its reason, together with the overall goal and previous subgoals, to generate a new subgoal or keep the existing one. In this way, verbose reason are distilled into concise subgoals.
\begin{equation}
G^t_{sub}= f_{PlanAgent}(G,G^t_p,G^{t-1}_{sub},I^t_{sel}, I_{hist},R^t_{sel})
\label{con:SubGoal}
\end{equation}
\textbf{Action Selection.}
In the fine-grained action selection stage, we use the depth map to identify traversable regions by treating areas whose height difference from the ground is below a threshold $ T_h $ as reachable. Starting from the agent’s current position, we then compute the farthest reachable point within the traversable region, and sample additional candidate reachable positions to the left and right of that point at a fixed angular interval.
To prevent candidates from being too close to boundaries, potentially causing collisions or out-of-bounds movements, we shrink each candidate position inward by a ratio $ Ratio_{shrink} $, yielding the final list of candidate reachable positions. We annotate these candidates and their indices on the image to obtain $ I_{ann} $.
We denote $ G, \, G^{t}_p, \, G^{t}_{sub} $ collectively as \(G_{all}\).
Next, we feed $ I_{ann} $ with the $ G_{all} $ into the agent, which selects the optimal candidate position. Finally, via coordinate transformation, we convert the selected position into a polar-coordinate representation, which is used as the expected action.
\begin{equation}
\begin{gathered}
A^e_t= f_{coord}(f_{ActionAgent}(G^t_{all},I^t_{ann}))
\label{con:ActionAgent}
\end{gathered}
\end{equation}

\textbf{Stop Decision.}
If the goal is observed, we treat the position in the annotated map that is closest to the goal as the goal location; otherwise, the goal location is left undefined. After each action execution, we further check whether the distance between the agent’s current position and the goal location is below a threshold $ T_s $. If it is, the agent stops at the current position and terminates the episode. In addition, when the number of executed steps reaches a predefined maximum $ T_{max} $, the agent also stops and ends the episode to avoid meaningless repeated exploration.
\begin{equation}
P_t^\star =
\begin{cases}
\displaystyle f_{StopAgent}(G^t_{all},I^t_{ann}), & if \, goal \, find,\\
\varnothing, & otherwise.
\end{cases}
\end{equation}
\begin{equation}
Stop_t =
\begin{cases}
1, & if \, (|P_t - P_t^\star | < T_s \lor\ t \ge T_{max}),\\
0, & otherwise.
\end{cases}
\end{equation}

\subsection{Review}
Most existing TF VLN methods rely on a single, static forward decision-making loop, which typically performs well only under relatively ideal conditions. In complex environments or anomalous scenarios, such approaches lack adaptability and are prone to error accumulation, which can progressively amplify decision bias.
Therefore, we propose a Review Module: a verification mechanism that does not rely on external supervision, but instead leverages self-feedback signals generated during execution to continuously check whether the current state is consistent with the expected state. When the review module detects no issues, the system proceeds normally into the execute module; when an anomaly is detected, the system revises the phase goal and switches to the corresponding error-correction mode.
This design minimizes interference with the normal pipeline, making the overall process smoother and more robust.
It comprises two mechanisms: a fine-grained two-step review and a coarse-grained multi-step review.

\textbf{Two-step Review.} Beyond the observation at time step $ t $, the change (delta) between time steps $ t $ and $ t-1 $ is also critical. Accordingly, we cache the observation image at $ t-1 $ in the memory module and use the images at $ t $ and $ t-1 $ jointly as inputs to the two-step review.
We first check whether the visual delta between the two time steps is below a threshold, or in other words, whether their similarity exceeds $ T_{sim} $. We then verify whether the positional change across the two time steps remains unchanged for two consecutive steps (to reduce false positives). If either condition holds, we conclude that the embodied agent is stuck due to an obstacle: the expected action in the previous step was not executed successfully.
In this case, the system enters an obstacle-skipping mode. The phase goal for the current step is set to “avoid the obstacle.” We exclude the obstructed direction (i.e., the previously chosen optimal direction $ D^{t-1}_{sel} $) and instead select a suboptimal alternative direction to re-explore, thereby restoring effective progress.
\begin{equation}
D^t_{sel}= \arg\max_{i\in \mathcal{D} \setminus \{ D^{t-1}_{sel} \}}\ {Score}^i_t
\label{con:Direction_sel}
\end{equation}

\begin{table*}[t]
\caption{\textcolor{black}{Main Results}}
\setlength{\tabcolsep}{5pt}  
\begin{tabular}{c|cc|cc|cc|cc|cc}
\hline
\multirow{2}{*}{Methods} & \multirow{2}{*}{TF} & \multirow{2}{*}{ZS} & \multicolumn{2}{c|}{MP3D}        & \multicolumn{2}{c|}{HM3D\_v0.1}   & \multicolumn{2}{c|}{HM3D\_v0.2}   & \multicolumn{2}{c}{HM3D\_OVON}  \\
                         &                              &           & SR             & SPL            & SR            & SPL            & SR            & SPL            & SR             & SPL            \\
\hline
ZSON \pub{NIPS} \cite{majumdar2022zson}        & \xmark      & \cmark             & 15.3              & 4.8              & 25.5          & 12.6           & -             & -              & -              & -              \\
PixNav \pub{ICRA} \cite{cai2024bridging}              & \xmark      & \cmark              & -              & -              & 37.9          & 20.5           & -             & -              & -              & -              \\
PSL \pub{ECCV} \cite{sun2024prioritized}                  & \xmark       & \cmark       & 18.9           & 6.4            & 42.4          & 19.2           & -             & -              & -              & -              \\
SGM \pub{CVPR} \cite{zhang2024imagine}             & \xmark            & \cmark              & 37.7           & 14.7           & \underline{60.2}          & 30.8           & -             & -              & -              & -              \\
VLFM \pub{ICRA} \cite{yokoyama2024vlfm}           & \xmark          & \cmark             & 36.4           & \underline{17.5}           & 52.5          & 30.4           & 62.6             & 31.0              & 35.2              & 19.6              \\
Uni-Navid \pub{arXiv} \cite{zhang2024uni}           & \xmark       & \cmark                & -              & -              & -             & -              & -             & -              & \underline{39.5}           & \underline{19.8}           \\
\hline
CoW \pub{CVPR} \cite{gadre2023cows}                  & \cmark       & \cmark              & 9.2            & 4.9            & -             & -              & -             & -              & -              & -              \\
ESC \pub{ICML} \cite{zhou2023esc}                & \cmark        & \cmark               & 28.7           & 14.2           & 39.2          & 22.3           & -             & -              & -              & -              \\
L3MVN \pub{IROS} \cite{yu2023l3mvn}              & \cmark      & \cmark               & -              & -              & 50.4          & 23.1           & 36.3             & 15.7              & -              & -              \\
VoroNav \pub{ICML} \cite{wu2024voronav}              & \cmark         & \cmark              & -              & -              & 42.0            & 26.0             & -             & -              & -              & -              \\
TopV-Nav \pub{arXiv} \cite{zhong2024topv}              & \cmark       & \cmark                & 35.2           & 16.4           & 53            & 29.8          & -             & -              & -              & -              \\
OpenFMNav \pub{ICLR} \cite{kuang2024openfmnav}            & \cmark          & \cmark     & -              & -              & 54.9          & 24.4           & -             & -              & -              & -              \\
SG-Nav \pub{NIPS} \cite{yin2024sg}            & \cmark          & \cmark     & 40.2              & 16.0              & 54.0          & 24.9           & 49.6             & 25.5              & -              & -              \\
VLMNav  \pub{NeuS} \cite{goetting2025end}          & \cmark      & \cmark                & -              & -              & 50.4          & 21.0           & -             & -              & -              & -              \\
UniGoal  \pub{CVPR} \cite{yin2025unigoal}          & \cmark      & \cmark                & 41.0              & 16.4              & 54.5          & 25.1           & -             & -              & -              & -              \\
InstructNav \pub{CoRL} \cite{long2025instructnav}            & \cmark          & \cmark     & -              & -              & 58.0          & 20.9           & -             & -              & -              & -              \\
MFNP \pub{arXiv} \cite{zhang2025multi}            & \cmark          & \cmark     & 41.1              & 15.4              & 58.3          & 26.7           & -             & -              & -              & -              \\
TopoNav \pub{arXiv} \cite{liu2025toponav}              & \cmark         & \cmark           &  \underline{45.5}    &  16.8     & 60.1          & \underline{34.6}          &  -    & -     & -              & -              \\
WMNav \pub{IROS} \cite{Nie2025WMNavIV}              & \cmark         & \cmark           & 45.4     &  17.2     & 58.1          & 31.2           &  \underline{72.2}    & \underline{33.3}     & -              & -              \\
TANGO \pub{CVPR} \cite{ziliotto2025tango}              & \cmark         & \cmark           & -    &  -     & -         & -          &  -    & -     & 35.5              & 19.5              \\
\textbf{MerNav}              & \cmark     & \cmark                      & \textbf{50.8} & \textbf{19.5}              & \textbf{68.0} & \textbf{36.9} & \textbf{74.8} & \textbf{37.6}                & \textbf{45.7} & \textbf{21.8}               \\
\hline
\end{tabular}
\label{tab:MainResults}
\end{table*}
\textbf{Multi-step Review.} 
An intuitive strategy is as follows: assuming the agent has target recognition capability, it systematically traverses all rooms, prioritizing those with a higher probability of containing the target. Once the current room has been fully explored, it proceeds to the next room in the traversal. If every room on the current floor has been traversed and the target is still not found, the agent transitions to another floor and continues the search. This “probability-guided approximate depth-first traversal” closely resembles how humans search for targets indoors.
To implement this strategy, we design a multi-step review mechanism with two components: exiting the room and exiting the floor.

\textit{Jump out of the Room}: We incorporate an exploration-area record map as part of the input, guiding the observation-analysis agent and the path-planning agent to determine whether the current room has been sufficiently explored. If exploration is confirmed complete, the agent should prioritize moving toward other unexplored areas and assign higher scores to directions leading to such areas, encouraging exploration of more valuable rooms.
\begin{equation}
G_p=
\begin{cases}
FindStairs, 
& if \; Flag_{jump}=1, \\
ReachNewFloor,
& if \; Flag_{stairs}=1,\\
\varnothing,
& if \; Flag_{floor}=1.
\end{cases}
\end{equation}
\textit{Jump out of the Floor}: When the current floor is detected to be fully explored, or when the accumulated step count reaches a preset threshold $ T_e $, the system triggers the floor-jump mode, in which case $ Flag_{exit}$ is set to true. The procedure is as follows: first, the phase goal $ G_p $ is switched to “find the stairs.” Next, once the stairs are found, $ Flag_{stairs} $ becomes true, the phase goal is updated to “reach a new floor,” with the subgoal constrained to “go up the stairs” or “go down the stairs.” Finally, upon arriving at the new floor, $ Flag_{floor} $ is set to true, the phase goal is cleared, and the agent continues searching for the goal $ G $ on the new floor.

\section{Experiments}
\subsection{Experimental Setup}
\textbf{Benchmarks.}
\underline{\textit{MP3D}}, an Object Goal Navigation (OGN) dataset was constructed based on MP3D scenes \cite{chang2017matterport3d}.
\underline{\textit{HM3D\_v0.1}} \cite{ramakrishnan2habitat} is designed for OGN based on HM3D scenes \cite{ramakrishnan2habitat}.
\underline{\textit{HM3D\_v0.2}} \cite{yadav2023habitat}, Compared to HM3D\_v0.1, it performs data cleaning.
\underline{\textit{HM3D\_OVON}} \cite{yokoyama2024hm3d} broadens the scope and semantic range of prior OGN benchmarks.

\textbf{Metrics.}
We use Success Rate (SR) and Success weighted by Path Length (SPL, can quantify the agent's trajectory efficiency) \cite{anderson2018evaluation} as evaluation metrics.

\textbf{Implementation Details.}
We follow the official configurations on MP3D, HM3D\_v0.1, and HM3D\_v0.2 \cite{batra2020objectnav}, and also on HM3D\_OVON \cite{yokoyama2024hm3d}. In addition, in HM3D\_OVON, the official setup does not restrict the camera pitch angle, we fix it at $-30^\circ$. The official settings of the maximum number of steps $ T_{max} $ is 500, to reduce cost, we limit $ T_{max} $ to 100.
\subsection{Main Results}
\textbf{Compared with all TF methods.}
As shown in Table \ref{tab:MainResults}, our method is training-free. Compared with all TF baselines, our approach achieves the best performance on both Success Rate and SPL across all 4 datasets. On average, the Success Rate improves by 7\%, with gains of 5\%, 8\%, 3\%, and 10\% on the four datasets, respectively; meanwhile, SPL improves by 3\% on average, with corresponding gains of 2\%, 2\%, 4\%, and 2\%.
Notably, the most pronounced improvements in Success Rate are observed on HM3D\_v0.1, the most commonly used dataset for the OGN task, and HM3D\_OVON, the newest and most challenging dataset, reaching 8\% (MerNav vs. TopoNav, etc.) and 10\% (MerNav vs. TANGO, etc.), respectively.

\textbf{Compared with all methods under the zero-shot setting.}
As shown in Table \ref{tab:MainResults}, our method operates in a zero-shot setting. Compared with all ZS baselines, it also achieves the best performance on both Success Rate and SPL across all 4 datasets, with an average improvement of 5\% in Success Rate and 3\% in SPL.
Notably, on HM3D\_v0.1 and HM3D\_OVON, the Success Rate increases by 8\% (MerNav vs. SGM, etc.) and 6\% (MerNav vs. Uni-Navid, etc.).
\begin{table}[h]
  \centering
  \caption{Comparison with the SFT method on the Open-Vocabulary Object Goal Navigation (HM3D\_OVON).}
  \label{tab:results}
  \setlength{\tabcolsep}{10pt}
  \begin{tabular}{lcccccc}
    \toprule
    \multirow{2}{*}{Methods} & \multirow{2}{*}{TF} & \multirow{2}{*}{ZS}
      & \multicolumn{2}{c}{MP3D} & \multicolumn{2}{c}{HM3D\_OVON} \\
      &  &  & SR & SPL & SR & SPL \\
    \midrule
    Habitat-Web \pub{CVPR} \cite{ramrakhya2022habitat}  & \xmark & \xmark &     31.6           & 8.5   & -            & - \\
    OVRL \pub{ICLR} \cite{yadav2023offline}  & \xmark & \xmark & 28.6              & 7.4    & -            & - \\
    T-Diff \pub{NIPS} \cite{yu2024trajectory} & \xmark & \xmark & 39.6              & 15.2    & -            & - \\
    DAgRL+OD \pub{IROS} \cite{yokoyama2024hm3d} & \xmark & \xmark & - & - & 37.1            & 19.9 \\
    VLFM \pub{ICRA}  \cite{yokoyama2024vlfm}    & \xmark & \xmark & - & - & 38.5            & 22.2 \\
    MTU3D \pub{ECCV} \cite{zhu2025move}    & \xmark & \xmark & - & - & 40.8            & 12.1 \\
    FiLM-Nav \pub{arXiv}  \cite{yokoyama2025film} & \xmark & \xmark & - & - & 40.8            & \textbf{24.4} \\
    Nav-R1 \pub{arXiv} \cite{liu2025nav} & \xmark & \xmark & - & - & 42.2            & 20.1 \\
    Nav-$ R^2 $ \pub{arXiv}  \cite{xiang2025nav} & \xmark & \xmark & - & - & 44.0            & 18.0 \\
    \hline
    MerNav                 & \cmark & \cmark & \textbf{50.8} & \textbf{19.5} & \textbf{45.7} & 21.8  \\
    \bottomrule
  \end{tabular}
  \label{tab:all}
\end{table}

\textbf{Compared with all methods.}
As a supplement to Table \ref{tab:MainResults}, Table \ref{tab:all} shows that our method achieves higher Success Rate (5\% and 2\%)  on MP3D and HM3D\_OVON than not only to all TF methods but also to all SFT methods. While maintaining the high generalization ability of TF methods, our method addresses the shortcomings of TF methods in terms of low success rate, achieving overall leading performance in both success rate and generalization ability.

Notably, although SFT methods typically perform well on in-domain data, their performance drops substantially under the zero-shot setting. we reproduce JanusVLN \cite{zeng2025janusvln} (a recent state-of-the-art method) and report the results in Table \ref{tab:Zero-Shot}. Compared with the version trained on the HM3D\_v0.1 training set (denoted JanusVLN*), directly transferring it to this dataset in a zero-shot manner without any training on HM3D\_v0.1 (denoted JanusVLN$ \ddagger $) leads to an approximately 11\% drop in SR and a 2\% drop in SPL.
In contrast, our method is inherently zero-shot, and thus is not affected by this issue, maintaining stable performance under cross-domain/zero-shot evaluation.
\begin{table}[h]
  \centering
  \caption{The Impact of the Zero-Shot Setting of Supervised Fine-Tuning methods on HM3D\_v0.1.}
  \label{tab:results}
  \setlength{\tabcolsep}{12pt}
  \begin{tabular}{lcccc}
    \toprule
    Methods & TF & ZS & SR & SPL \\
    \midrule
    JanusVLN* \pub{ICLR} \cite{zeng2025janusvln} & \xmark     & \xmark & 62.6  & 30.8 \\
    JanusVLN$ \ddagger $ \pub{ICLR} \cite{zeng2025janusvln}  & \xmark     & \cmark & 52.0  & 28.5 \\
    MerNav    & \cmark     & \cmark & \textbf{68.0} & \textbf{36.9} \\
    \bottomrule
  \end{tabular}
\label{tab:Zero-Shot}
\end{table}

\subsection{Ablation Experiment}
\textbf{Impact of Different Modules.}
We present the ablation results on the HM3D\_v0.1 dataset using a progressively additive, module-by-module setup, which makes the gains contributed by each module more transparent while effectively reducing computational cost. 
As shown in Table \ref{tab:ablation}, starting from the code base (WMNav), upgrading the backbone model from Gemini 1.5 Pro to Qwen3-vl-plus improves the success rate by 2\%. This result highlights one advantage of the TF approach: it can readily benefit from continuous improvements in backbone model capability.
After introducing the memory module, since it primarily serves to support subsequent modules, its contribution is reflected in later ablation settings. In addition, we report the result of adding only commonsense memory: SPL shows a slight improvement. Given that SPL is harder to improve than SR, this gain is meaningful.
With the review module further added, the success rate increases by another 6\%, including a 2\% improvement for “exiting the room” and a 4\% improvement for “exiting the floor.” Finally, after improving the pipeline in the Execute module, the success rate increases by approximately 2\% more, while SPL also rises by about 2\%.
Overall, these results indicate that each module provides stable and complementary performance gains, validating the necessity of the proposed design.
\begin{table}[h]
\caption{\textcolor{black}{Impact of Different Modules}}
\centering
\setlength{\tabcolsep}{12pt}
\begin{tabular}{llrr}
\toprule
Module & Component & SR & SPL \\
\midrule
Basic Pipeline & + Qwen3-vl-plus & 60.25 & 32.35 \\
Memory & + Common Sense & 60.10 & 32.42 \\
\multirow{2}{*}{Review} & + Jump out Room & 62.30 & 33.82 \\
 & + Jump out Floor & 66.30 & 34.59 \\
Execute & + Analysis, etc & \textbf{67.95} & \textbf{36.92} \\
\bottomrule
\end{tabular}
\label{tab:ablation}
\end{table}

\textbf{Impact of Different Foundation Models.}
The TF model can be viewed as a combination of a “foundation model + domain adaptation module,” where the capability ceiling of the foundation model has a substantial impact on the final performance. As shown in Table 3, we compare different foundation models. The results indicate that the latest GPT-5.2 model \cite{openai2025introducinggpt52}(released on Dec 11, 2025) demonstrates a significant capability improvement over Qwen3-vl-plus \cite{bai2025qwen3vltechnicalreport} (released on Sep 23, 2025), further boosting MerNav’s performance on HM3D\_v0.1 by 3\%. 
At this point, compared to the current best Zero-Shot baseline, the success rate increased by 11\%, reaching a “Acceptable” level (over 70\%). 
However, considering cost constraints, we still adopt Qwen3-vl-plus as the foundation model for the main experiments and all module ablation studies.
\begin{table}[h]
\caption{Performance of MerNav with different foundation models.}
\centering
\setlength{\tabcolsep}{12pt}
\begin{tabular}{llcc}
\toprule
Method & Foundation Model & SR & SPL \\
\midrule
\multirow{2}{*}{MerNav} & Qwen3-vl-plus & 68.0 & 36.9 \\
                        & GPT-5.2       & \textbf{70.9}  & \textbf{39.7} \\
\bottomrule
\end{tabular}
\label{tab:mernav_foundation}
\end{table}

\subsection{Case Study}
\textbf{Interpretability.}
The following are examples of the observation analysis agent's output.
In the analysis of the 240° view, the model explicitly responds to the injected common-sense knowledge (“Please note that your body width is approximately 0.4 meters...”), indicating that common-sense injection encourages the model to provide clearer evidence along its reasoning pathway, thereby improving the transparency and interpretability of the decision-making process.
In the analysis of the 300° view, the model jointly considers image information from the local, global, and historical perspectives, demonstrating that these three sources play distinct yet complementary roles during the analysis. This suggests that multi-dimensional information fusion helps the model produce more reliable and higher-quality outputs in a more interpretable manner.
\begin{tcolorbox}[
  enhanced,
  breakable,
  colback=white,
  colframe=white,
  borderline west={2pt}{0pt}{black!30},
  left=6pt,right=6pt,top=4pt,bottom=4pt
]
\ttfamily\small
'240': \{'score': 6, 'reason': '...Since the direction aligns with an unexplored path and there’s no obstacle blocking forward movement \textbf{(width >0.4m)}, it’s moderately promising...'\}, \\
'300': \{'score': 3, 'reason': '\textbf{The view at 300 degrees} shows a partial wall and a large leather sofa, with no sign of a bed. \textbf{The panoramic view }confirms this is part of a living area, and \textbf{the top-down map} indicates that while the immediate vicinity (gray) has been explored...'\}
\end{tcolorbox}

\subsection{Humanoid Robot Real-world Experiment}
\begin{figure*}[t]
  \centering
  \includegraphics[width=0.95\linewidth]{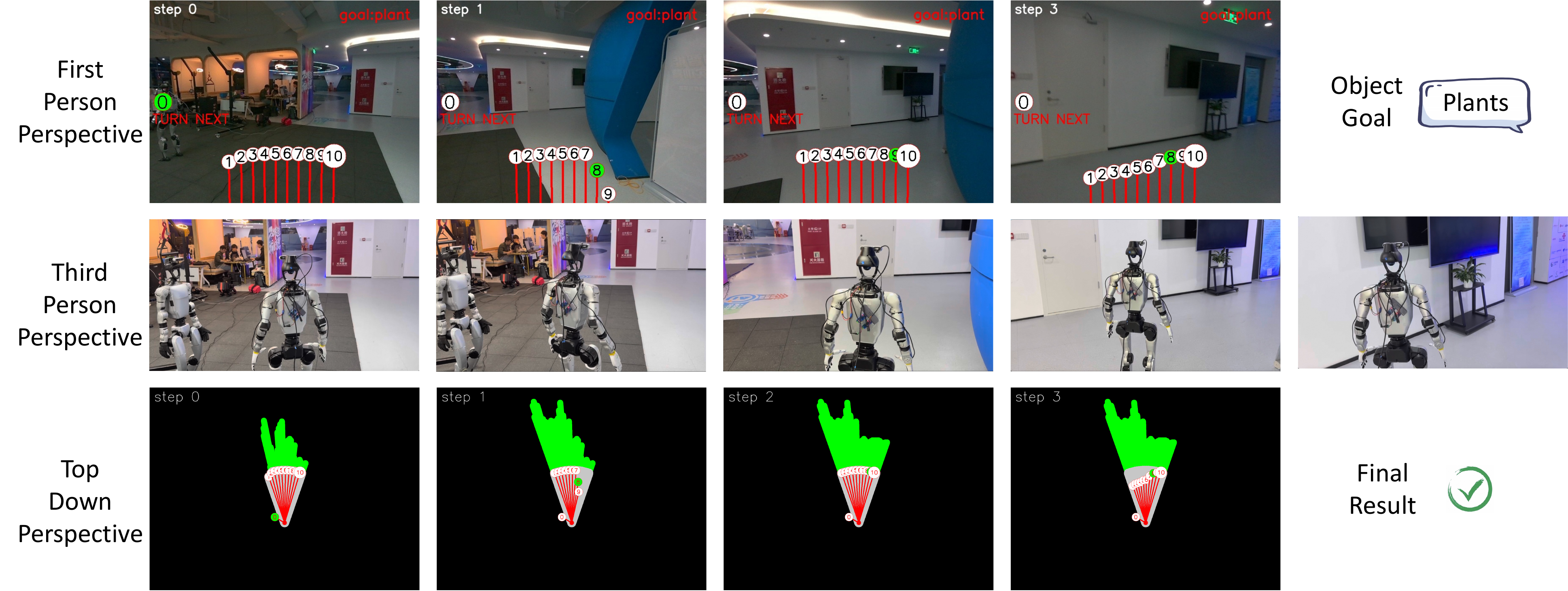}
  \caption{Humanoid Robot Real Case 1, Object Goal: Plants.}
  \label{fig:Plants}
\end{figure*}

\begin{figure*}[t]
  \centering
  \includegraphics[width=0.95\linewidth]{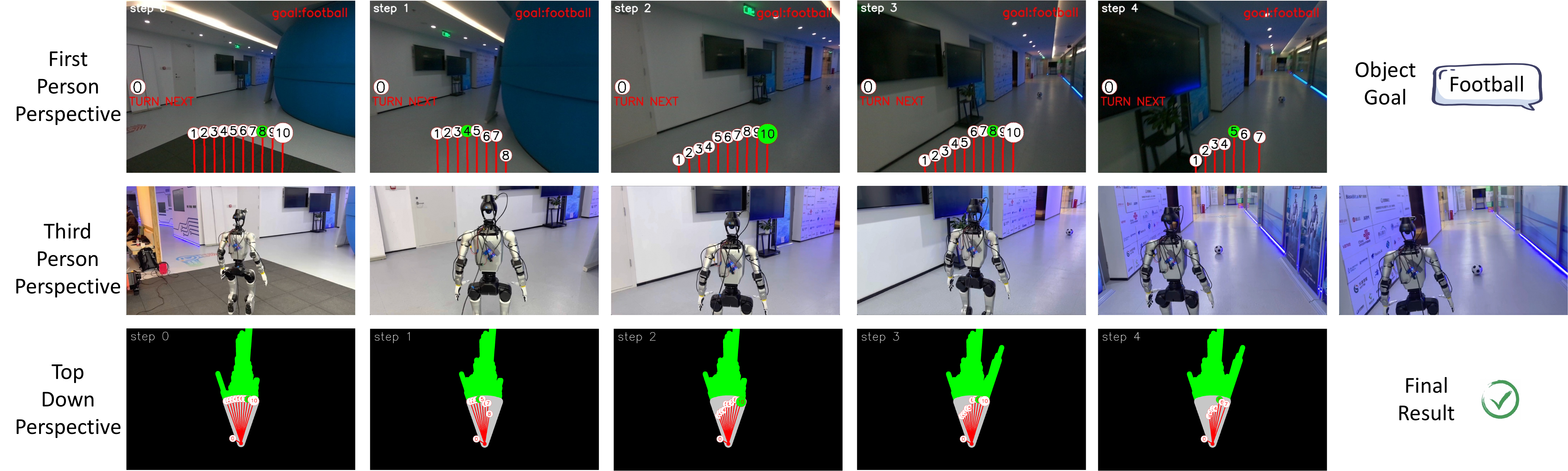}
  \caption{Humanoid Robot Real Case 2, Object Goal: Football.}
  \label{fig:football}
\end{figure*}

We deployed the MerNav model on the Yushu G1 humanoid robot, as shown in Figures \ref{fig:Plants} and \ref{fig:football}, presenting two cases with target objects: a plant and a soccer ball. The first row in the figure shows images from the robot's first-person perspective, with the model's decision for the next step location overlaid. The second row shows the third-person perspective images at the corresponding moments, including the image of the final stop position. The third row presents top-down views, demonstrating the process of the model constructing a map from depth maps.

Since the humanoid robot only has a front-facing camera (while robotic dogs typically have cameras at multiple angles), we did not capture panoramic images from multiple directions. Instead, the model determines whether the current direction is worth exploring. If it is not, the robot rotates 60 degrees clockwise. If it is worth exploring, the model selects the next step location. Detailed content can be found at https://qidekang.github.io/MerNav.github.io/

\section{Related Work}
\textbf{Supervised Fine-Tuning VLN.}
These methods leverage task-specific data to fine-tune multimodal foundation models or train policy models, and typically achieve high success rates in in-distribution settings. Include (1) end-to-end approaches that directly learn policies mapping observations to actions, such as \cite{ramrakhya2022habitat}, \cite{yadav2023offline}, and \cite{yokoyama2024hm3d}, and (2) methods that learn spatial representations (explicit maps or implicit spatial memory) for planning to improve stability, such as \cite{zhang2024imagine}, \cite{yu2024trajectory}, and \cite{zhu2025move}.
However, them require large amounts of training data, and their performance is strongly coupled with the training data distribution, often degrading noticeably under zero-shot evaluation or cross-dataset transfer.
In contrast, our method does not rely on any training data or training process, exhibits stronger generalization, and achieves success rates higher than most SFT-based VLN approaches.

\textbf{Training-Free VLN.}
These methods aim to preserve the general-purpose capability of the foundation model to maintain generalization, while improving success rates through strategies such as (1) rule-based constraints and heuristic search (e.g., \cite{zhou2023esc}, \cite{gadre2023cows}, \cite{goetting2025end}), (2) scene-graph and topology-graph construction (e.g., \cite{yin2024sg}, \cite{liu2025toponav}, \cite{wu2024voronav}), and (3) multi-stage task decomposition (e.g., \cite{yu2023l3mvn}, \cite{zhang2025multi}, \cite{ziliotto2025tango}). 
Nevertheless, being constrained by a single-round, static decision–action loop, they often struggle to adapt to complex scenarios.
Inspired by collaboration mechanisms in the human brain and human collectives, we build a memory–execute–review framework that enhances performance in both routine and complex situations, yielding substantial improvements in success rate and generalization.

\section{Conclusion}
VLN is one of the key challenges that embodied intelligence urgently needs to tackle, yet the overall performance of existing methods remains far from satisfactory.
Inspired by collaboration mechanisms in the human brain and human collectives, we propose the highly generalizable MerNav framework, composed of three modules: memory, execute, and review, improving the model's memory capability and its ability to handle routine and anomalous situations. it achieves comprehensive leading performance across multiple datasets, method types, and evaluation metrics.
We validate our framework on the OGN task. Across 4 datasets, our approach improves the the success rate by an average of 7\%, and 5\% compared to all methods under the TF and Zero-Shot setting, respectively. On the most commonly used HM3D\_v0.1 benchmark and the newer, more challenging open-vocabulary dataset HM3D\_OVON, the Success Rate increases by 8\% and 6\% in ZS setting, respectively.
Moreover, on MP3D and HM3D\_OVON, our method achieves higher Success Rate (5\% and 2\%) than not only all TF but also all SFT methods, demonstrating comprehensive advantages in both Success Rate and generalization.
In addition, adopting a more advanced foundation model can further boost Success Rate and SPL by approximately 3\% each, bringing the Success Rate to a "Acceptable" level ($ > $70\%).
Overall, these results demonstrate MerNav’s excellent Success Rate and generalization ability, highlight the substantial potential of TF VLN, and provide new momentum for VLN research and applications, especially for zero-shot VLN.

\bibliographystyle{plainnat}
\bibliography{references}  

@online{MorganStanley2025Humanoids5T,
  author       = {{Morgan Stanley Research}},
  title        = {Humanoids: A \$5 Trillion Market},
  year         = {2025},
  month        = may,
  day          = {14},
  url          = {https://www.morganstanley.com/insights/articles/humanoid-robot-market-5-trillion-by-2050},
  urldate      = {2026-01-19},
  organization = {Morgan Stanley},
  note         = {Morgan Stanley Insights (Research)}
}

@article{tulving1998episodic,
  title={Episodic and declarative memory: role of the hippocampus},
  author={Tulving, Endel and Markowitsch, Hans J},
  journal={Hippocampus},
  volume={8},
  number={3},
  pages={198--204},
  year={1998},
  publisher={Wiley Online Library}
}

@article{barbey2013dorsolateral,
  title={Dorsolateral prefrontal contributions to human working memory},
  author={Barbey, Aron K and Koenigs, Michael and Grafman, Jordan},
  journal={cortex},
  volume={49},
  number={5},
  pages={1195--1205},
  year={2013},
  publisher={Elsevier}
}

@article{raichle2015brain,
  title={The brain's default mode network},
  author={Raichle, Marcus E},
  journal={Annual review of neuroscience},
  volume={38},
  number={1},
  pages={433--447},
  year={2015},
  publisher={Annual Reviews}
}

@article{marek2018frontoparietal,
  title={The frontoparietal network: function, electrophysiology, and importance of individual precision mapping},
  author={Marek, Scott and Dosenbach, Nico UF},
  journal={Dialogues in clinical neuroscience},
  volume={20},
  number={2},
  pages={133--140},
  year={2018},
  publisher={Taylor \& Francis}
}

@article{yin2006role,
  title={The role of the basal ganglia in habit formation},
  author={Yin, Henry H and Knowlton, Barbara J},
  journal={Nature reviews neuroscience},
  volume={7},
  number={6},
  pages={464--476},
  year={2006},
  publisher={Nature Publishing Group UK London}
}

@article{schneider1977controlled,
  title={Controlled and automatic human information processing: I. Detection, search, and attention.},
  author={Schneider, Walter and Shiffrin, Richard M},
  journal={Psychological review},
  volume={84},
  number={1},
  pages={1},
  year={1977},
  publisher={American Psychological Association}
}

@article{swick2002dissociation,
  title={Dissociation between conflict detection and error monitoring in the human anterior cingulate cortex},
  author={Diane Swick and {{And U. Turken}}},
  journal={Proceedings of the National Academy of Sciences},
  volume={99},
  number={25},
  pages={16354--16359},
  year={2002},
  publisher={National Academy of Sciences}
}

@article{ham2013cognitive,
  title={Cognitive control and the salience network: an investigation of error processing and effective connectivity},
  author={Ham, Timothy and Leff, Alex and de Boissezon, Xavier and Joffe, Anna and Sharp, David J},
  journal={Journal of Neuroscience},
  volume={33},
  number={16},
  pages={7091--7098},
  year={2013},
  publisher={Society for Neuroscience}
}

@article{wu2025human,
  title={From human memory to ai memory: A survey on memory mechanisms in the era of llms},
  author={Wu, Yaxiong and Liang, Sheng and Zhang, Chen and Wang, Yichao and Zhang, Yongyue and Guo, Huifeng and Tang, Ruiming and Liu, Yong},
  journal={arXiv preprint arXiv:2504.15965},
  year={2025}
}

@article{lechner1999100,
  title={100 years of consolidation—remembering M{\"u}ller and Pilzecker},
  author={Lechner, Hilde A and Squire, Larry R and Byrne, John H},
  journal={Learning \& memory},
  volume={6},
  number={2},
  pages={77--87},
  year={1999},
  publisher={Cold Spring Harbor Lab}
}

@article{roselli2021making,
  title={The making of long-lasting memories: a fruit fly perspective},
  author={Roselli, Camilla and Ramaswami, Mani and Boto, Tamara and Cervantes-Sandoval, Isaac},
  journal={Frontiers in behavioral neuroscience},
  volume={15},
  pages={662129},
  year={2021},
  publisher={Frontiers Media SA}
}

@article{akhtar2018common,
  title={The ‘common sense’memory belief system and its implications},
  author={Akhtar, Shazia and Justice, Lucy V and Knott, Lauren and Kibowski, Fraenze and Conway, Martin A},
  journal={The International Journal of Evidence \& Proof},
  volume={22},
  number={3},
  pages={289--304},
  year={2018},
  publisher={SAGE Publications Sage UK: London, England}
}

@inproceedings{ramakrishnan2habitat,
  title={Habitat-Matterport 3D Dataset (HM3D): 1000 Large-scale 3D Environments for Embodied AI},
  author={Ramakrishnan, Santhosh Kumar and Gokaslan, Aaron and Wijmans, Erik and Maksymets, Oleksandr and Clegg, Alexander and Turner, John M and Undersander, Eric and Galuba, Wojciech and Westbury, Andrew and Chang, Angel X and others},
  booktitle={Thirty-fifth Conference on Neural Information Processing Systems Datasets and Benchmarks Track (Round 2)},
  year={2021}
}

@inproceedings{yadav2023habitat,
  title={Habitat-matterport 3d semantics dataset},
  author={Yadav, Karmesh and Ramrakhya, Ram and Ramakrishnan, Santhosh Kumar and Gervet, Theo and Turner, John and Gokaslan, Aaron and Maestre, Noah and Chang, Angel Xuan and Batra, Dhruv and Savva, Manolis and others},
  booktitle={Proceedings of the IEEE/CVF Conference on Computer Vision and Pattern Recognition},
  pages={4927--4936},
  year={2023}
}

@inproceedings{yokoyama2024hm3d,
  title={Hm3d-ovon: A dataset and benchmark for open-vocabulary object goal navigation},
  author={Yokoyama, Naoki and Ramrakhya, Ram and Das, Abhishek and Batra, Dhruv and Ha, Sehoon},
  booktitle={2024 IEEE/RSJ International Conference on Intelligent Robots and Systems (IROS)},
  pages={5543--5550},
  year={2024},
  organization={IEEE}
}

@article{chang2017matterport3d,
  title={Matterport3d: Learning from rgb-d data in indoor environments},
  author={Chang, Angel and Dai, Angela and Funkhouser, Thomas and Halber, Maciej and Niessner, Matthias and Savva, Manolis and Song, Shuran and Zeng, Andy and Zhang, Yinda},
  journal={arXiv preprint arXiv:1709.06158},
  year={2017}
}

@misc{habitatChallenge2021,
  title        = {Habitat Challenge 2021},
  author       = {{AI Habitat}},
  year         = {2021},
  howpublished = {\url{https://aihabitat.org/challenge/2021/}},
  note         = {Accessed: 2026-01-26}
}

@article{batra2020objectnav,
  title={Objectnav revisited: On evaluation of embodied agents navigating to objects},
  author={Batra, Dhruv and Gokaslan, Aaron and Kembhavi, Aniruddha and Maksymets, Oleksandr and Mottaghi, Roozbeh and Savva, Manolis and Toshev, Alexander and Wijmans, Erik},
  journal={arXiv preprint arXiv:2006.13171},
  year={2020}
}

@misc{habitatchallenge2022,
  title         =     {Habitat Challenge 2022},
  author        =     {Karmesh Yadav and Santhosh Kumar Ramakrishnan and John Turner and Aaron Gokaslan and Oleksandr Maksymets and Rishabh Jain and Ram Ramrakhya and Angel X Chang and Alexander Clegg and Manolis Savva and Eric Undersander and Devendra Singh Chaplot and Dhruv Batra},
  howpublished  =     {\url{https://aihabitat.org/challenge/2022/}},
  year          =     {2022}
}

@misc{habitatchallenge2023,
  title         =     {Habitat Challenge 2023},
  author        =     {Karmesh Yadav and Jacob Krantz and Ram Ramrakhya and Santhosh Kumar Ramakrishnan and Jimmy Yang and Austin Wang and John Turner and Aaron Gokaslan and Vincent-Pierre Berges and Roozbeh Mootaghi and Oleksandr Maksymets and Angel X Chang and Manolis Savva and Alexander Clegg and Devendra Singh Chaplot and Dhruv Batra},
  howpublished  =     {\url{https://aihabitat.org/challenge/2023/}},
  year          =     {2023}
}

@article{anderson2018evaluation,
  title={On evaluation of embodied navigation agents},
  author={Anderson, Peter and Chang, Angel and Chaplot, Devendra Singh and Dosovitskiy, Alexey and Gupta, Saurabh and Koltun, Vladlen and Kosecka, Jana and Malik, Jitendra and Mottaghi, Roozbeh and Savva, Manolis and others},
  journal={arXiv preprint arXiv:1807.06757},
  year={2018}
}

@online{openai2025introducinggpt52,
  author       = {{OpenAI}},
  title        = {Introducing GPT-5.2},
  year         = {2025},
  month        = dec,
  day          = {11},
  url          = {https://openai.com/index/introducing-gpt-5-2/},
  urldate      = {2026-01-26}
}

@misc{bai2025qwen3vltechnicalreport,
      title={Qwen3-VL Technical Report}, 
      author={Shuai Bai and Yuxuan Cai and Ruizhe Chen and Keqin Chen and Xionghui Chen and Zesen Cheng and Lianghao Deng and Wei Ding and Chang Gao and Chunjiang Ge and Wenbin Ge and Zhifang Guo and Qidong Huang and Jie Huang and Fei Huang and Binyuan Hui and Shutong Jiang and Zhaohai Li and Mingsheng Li and Mei Li and Kaixin Li and Zicheng Lin and Junyang Lin and Xuejing Liu and Jiawei Liu and Chenglong Liu and Yang Liu and Dayiheng Liu and Shixuan Liu and Dunjie Lu and Ruilin Luo and Chenxu Lv and Rui Men and Lingchen Meng and Xuancheng Ren and Xingzhang Ren and Sibo Song and Yuchong Sun and Jun Tang and Jianhong Tu and Jianqiang Wan and Peng Wang and Pengfei Wang and Qiuyue Wang and Yuxuan Wang and Tianbao Xie and Yiheng Xu and Haiyang Xu and Jin Xu and Zhibo Yang and Mingkun Yang and Jianxin Yang and An Yang and Bowen Yu and Fei Zhang and Hang Zhang and Xi Zhang and Bo Zheng and Humen Zhong and Jingren Zhou and Fan Zhou and Jing Zhou and Yuanzhi Zhu and Ke Zhu},
      year={2025},
      eprint={2511.21631},
      archivePrefix={arXiv},
      primaryClass={cs.CV},
      url={https://arxiv.org/abs/2511.21631}, 
}

@article{yokoyama2025film,
  title={FiLM-Nav: Efficient and Generalizable Navigation via VLM Fine-tuning},
  author={Yokoyama, Naoki and Ha, Sehoon},
  journal={arXiv preprint arXiv:2509.16445},
  year={2025}
}

@inproceedings{ziliotto2025tango,
  title={TANGO: training-free embodied AI agents for open-world tasks},
  author={Ziliotto, Filippo and Campari, Tommaso and Serafini, Luciano and Ballan, Lamberto},
  booktitle={Proceedings of the Computer Vision and Pattern Recognition Conference},
  pages={24603--24613},
  year={2025}
}

@inproceedings{zhang2025multi,
  title={Multi-floor zero-shot object navigation policy},
  author={Zhang, Lingfeng and Wang, Hao and Xiao, Erjia and Zhang, Xinyao and Zhang, Qiang and Jiang, Zixuan and Xu, Renjing},
  booktitle={2025 IEEE International Conference on Robotics and Automation (ICRA)},
  pages={6416--6422},
  year={2025},
  organization={IEEE}
}

@inproceedings{long2025instructnav,
  title={InstructNav: Zero-shot System for Generic Instruction Navigation in Unexplored Environment},
  author={Long, Yuxing and Cai, Wenzhe and Wang, Hongcheng and Zhan, Guanqi and Dong, Hao},
  booktitle={Conference on Robot Learning},
  pages={2049--2060},
  year={2025},
  organization={PMLR}
}

@article{liu2025toponav,
  title={Toponav: Topological graphs as a key enabler for advanced object navigation},
  author={Liu, Peiran and Zhang, Qiang and Peng, Daojie and Zhang, Lingfeng and Qin, Yihao and Zhou, Hang and Ma, Jun and Xu, Renjing and Ji, Yiding},
  journal={arXiv preprint arXiv:2509.01364},
  year={2025}
}

@inproceedings{yin2025unigoal,
  title={Unigoal: Towards universal zero-shot goal-oriented navigation},
  author={Yin, Hang and Xu, Xiuwei and Zhao, Linqing and Wang, Ziwei and Zhou, Jie and Lu, Jiwen},
  booktitle={Proceedings of the Computer Vision and Pattern Recognition Conference},
  pages={19057--19066},
  year={2025}
}

@article{yin2024sg,
  title={Sg-nav: Online 3d scene graph prompting for llm-based zero-shot object navigation},
  author={Yin, Hang and Xu, Xiuwei and Wu, Zhenyu and Zhou, Jie and Lu, Jiwen},
  journal={Advances in neural information processing systems},
  volume={37},
  pages={5285--5307},
  year={2024}
}

@article{yu2024trajectory,
  title={Trajectory diffusion for objectgoal navigation},
  author={Yu, Xinyao and Zhang, Sixian and Song, Xinhang and Qin, Xiaorong and Jiang, Shuqiang},
  journal={Advances in Neural Information Processing Systems},
  volume={37},
  pages={110388--110411},
  year={2024}
}

@article{liu2025nav,
  title={Nav-r1: Reasoning and navigation in embodied scenes},
  author={Liu, Qingxiang and Huang, Ting and Zhang, Zeyu and Tang, Hao},
  journal={arXiv preprint arXiv:2509.10884},
  year={2025}
}

@article{xiang2025nav,
  title={Nav-$ R^2$ Dual-Relation Reasoning for Generalizable Open-Vocabulary Object-Goal Navigation},
  author={Xiang, Wentao and Zhang, Haokang and Yang, Tianhang and Chu, Zedong and Chu, Ruihang and Xie, Shichao and Yuan, Yujian and Sun, Jian and Gu, Zhining and Wang, Junjie and others},
  journal={arXiv preprint arXiv:2512.02400},
  year={2025}
}

@inproceedings{ramrakhya2022habitat,
  title={Habitat-web: Learning embodied object-search strategies from human demonstrations at scale},
  author={Ramrakhya, Ram and Undersander, Eric and Batra, Dhruv and Das, Abhishek},
  booktitle={Proceedings of the IEEE/CVF conference on computer vision and pattern recognition},
  pages={5173--5183},
  year={2022}
}

@inproceedings{yadav2023offline,
  title={Offline visual representation learning for embodied navigation},
  author={Yadav, Karmesh and Ramrakhya, Ram and Majumdar, Arjun and Berges, Vincent-Pierre and Kuhar, Sachit and Batra, Dhruv and Baevski, Alexei and Maksymets, Oleksandr},
  booktitle={Workshop on Reincarnating Reinforcement Learning at ICLR 2023},
  year={2023}
}

@article{zeng2025janusvln,
  title={Janusvln: Decoupling semantics and spatiality with dual implicit memory for vision-language navigation},
  author={Zeng, Shuang and Qi, Dekang and Chang, Xinyuan and Xiong, Feng and Xie, Shichao and Wu, Xiaolong and Liang, Shiyi and Xu, Mu and Wei, Xing},
  journal={arXiv preprint arXiv:2509.22548},
  year={2025}
}

@article{majumdar2022zson,
  title={Zson: Zero-shot object-goal navigation using multimodal goal embeddings},
  author={Majumdar, Arjun and Aggarwal, Gunjan and Devnani, Bhavika and Hoffman, Judy and Batra, Dhruv},
  journal={Advances in Neural Information Processing Systems},
  volume={35},
  pages={32340--32352},
  year={2022}
}

@inproceedings{cai2024bridging,
  title={Bridging zero-shot object navigation and foundation models through pixel-guided navigation skill},
  author={Cai, Wenzhe and Huang, Siyuan and Cheng, Guangran and Long, Yuxing and Gao, Peng and Sun, Changyin and Dong, Hao},
  booktitle={2024 IEEE International Conference on Robotics and Automation (ICRA)},
  pages={5228--5234},
  year={2024},
  organization={IEEE}
}

@inproceedings{sun2024prioritized,
  title={Prioritized semantic learning for zero-shot instance navigation},
  author={Sun, Xinyu and Liu, Lizhao and Zhi, Hongyan and Qiu, Ronghe and Liang, Junwei},
  booktitle={European Conference on Computer Vision},
  pages={161--178},
  year={2024},
  organization={Springer}
}

@inproceedings{yokoyama2024vlfm,
  title={Vlfm: Vision-language frontier maps for zero-shot semantic navigation},
  author={Yokoyama, Naoki and Ha, Sehoon and Batra, Dhruv and Wang, Jiuguang and Bucher, Bernadette},
  booktitle={2024 IEEE International Conference on Robotics and Automation (ICRA)},
  pages={42--48},
  year={2024},
  organization={IEEE}
}

@inproceedings{zhang2024imagine,
  title={Imagine before go: Self-supervised generative map for object goal navigation},
  author={Zhang, Sixian and Yu, Xinyao and Song, Xinhang and Wang, Xiaohan and Jiang, Shuqiang},
  booktitle={Proceedings of the IEEE/CVF Conference on Computer Vision and Pattern Recognition},
  pages={16414--16425},
  year={2024}
}

@article{zhang2024uni,
  title={Uni-navid: A video-based vision-language-action model for unifying embodied navigation tasks},
  author={Zhang, Jiazhao and Wang, Kunyu and Wang, Shaoan and Li, Minghan and Liu, Haoran and Wei, Songlin and Wang, Zhongyuan and Zhang, Zhizheng and Wang, He},
  journal={arXiv preprint arXiv:2412.06224},
  year={2024}
}

@inproceedings{zhu2025move,
  title={Move to understand a 3d scene: Bridging visual grounding and exploration for efficient and versatile embodied navigation},
  author={Zhu, Ziyu and Wang, Xilin and Li, Yixuan and Zhang, Zhuofan and Ma, Xiaojian and Chen, Yixin and Jia, Baoxiong and Liang, Wei and Yu, Qian and Deng, Zhidong and others},
  booktitle={Proceedings of the IEEE/CVF International Conference on Computer Vision},
  pages={8120--8132},
  year={2025}
}

@inproceedings{gadre2023cows,
  title={Cows on pasture: Baselines and benchmarks for language-driven zero-shot object navigation},
  author={Gadre, Samir Yitzhak and Wortsman, Mitchell and Ilharco, Gabriel and Schmidt, Ludwig and Song, Shuran},
  booktitle={Proceedings of the IEEE/CVF Conference on Computer Vision and Pattern Recognition},
  pages={23171--23181},
  year={2023}
}

@inproceedings{zhou2023esc,
  title={Esc: Exploration with soft commonsense constraints for zero-shot object navigation},
  author={Zhou, Kaiwen and Zheng, Kaizhi and Pryor, Connor and Shen, Yilin and Jin, Hongxia and Getoor, Lise and Wang, Xin Eric},
  booktitle={International Conference on Machine Learning},
  pages={42829--42842},
  year={2023},
  organization={PMLR}
}

@inproceedings{wu2024voronav,
  title={VoroNav: voronoi-based zero-shot object navigation with large language model},
  author={Wu, Pengying and Mu, Yao and Wu, Bingxian and Hou, Yi and Ma, Ji and Zhang, Shanghang and Liu, Chang},
  booktitle={Proceedings of the 41st International Conference on Machine Learning},
  pages={53737--53775},
  year={2024}
}

@inproceedings{yu2023l3mvn,
  title={L3mvn: Leveraging large language models for visual target navigation},
  author={Yu, Bangguo and Kasaei, Hamidreza and Cao, Ming},
  booktitle={2023 IEEE/RSJ International Conference on Intelligent Robots and Systems (IROS)},
  pages={3554--3560},
  year={2023},
  organization={IEEE}
}

@inproceedings{goetting2025end,
  title={End-to-End Navigation with Vision-Language Models: Transforming Spatial Reasoning into Question-Answering},
  author={Goetting, Dylan and Singh, Himanshu Gaurav and Loquercio, Antonio},
  booktitle = {Proceedings of the International Conference on Neuro-symbolic Systems},
  series    = {Proceedings of Machine Learning Research},
  volume    = {288},
  pages     = {22--35},
  year      = {2025},
  publisher = {PMLR}
}

@inproceedings{kuang2024openfmnav,
  title={OpenFMNav: Towards Open-Set Zero-Shot Object Navigation via Vision-Language Foundation Models},
  author={Kuang, Yuxuan and Lin, Hai and Jiang, Meng},
  booktitle={ICLR 2024 Workshop on Large Language Model (LLM) Agents},
  year={2024}
}

@article{zhong2024topv,
  title={Topv-nav: Unlocking the top-view spatial reasoning potential of mllm for zero-shot object navigation},
  author={Zhong, Linqing and Gao, Chen and Ding, Zihan and Liao, Yue and Ma, Huimin and Zhang, Shifeng and Zhou, Xu and Liu, Si},
  journal={arXiv preprint arXiv:2411.16425},
  year={2024}
}

@article{Nie2025WMNavIV,
  title={WMNav: Integrating Vision-Language Models into World Models for Object Goal Navigation},
  author={Dujun Nie and Xianda Guo and Yiqun Duan and Ruijun Zhang and Long Chen},
  journal={2025 IEEE/RSJ International Conference on Intelligent Robots and Systems (IROS)},
  year={2025},
  pages={2392-2399},
  url={https://api.semanticscholar.org/CorpusID:276776282}
}

\section{Appendices}
\subsection{Reason for the Module Name}
It’s worth noting that review and reflection differ in emphasis: reflection tends to come from a self-centered perspective, whereas review is more like looking from an “observer” standpoint, or from a higher-level vantage point of your own.
There is also a distinction between action and execution: action usually refers to a single, concrete act, while execution is a more systematic and comprehensive process. By analogy to a CEO’s (Chief Executive Officer) responsibilities, execution often requires coordinating resources, aligning efforts, and turning work into delivered results, etc.

\subsection{VLN Task Categorization and Selection}
VLN tasks are typically formulated under several settings, such as Object Goal Navigation, Instruction-following Navigation, Point Goal Navigation, and Image Goal Navigation, which require different types of inputs including a target category name, fine-grained step-by-step instructions, a point coordinate, or a target image. In this paper, we adopt the Object Goal Navigation setting because it better aligns with the natural interaction logic between humans and robots. The task can be formalized as follows.

\subsection{Details of Datasets and Metrics}
\textbf{Benchmarks.}
HM3D is a large-scale dataset of 1,000 building-scale 3D reconstructions of real-world locations. \cite{ramakrishnan2habitat}.

\underline{\textit{HM3D\_v0.1}} (HM3D-Semantics v0.1) \cite{ramakrishnan2habitat} was used in the Habitat Challenge 2022 \cite{habitatchallenge2022}. It is designed for Object Goal Navigation and includes 6 object-goal categories: chair, couch, potted plant, bed, toilet, and tv. Its validation set contains 2000 episodes.

\underline{\textit{HM3D\_v0.2}} (HM3D-Semantics v0.2) \cite{yadav2023habitat} was used in the Habitat Challenge 2023 \cite{habitatchallenge2023}. Compared to HM3D\_v0.1, it performs data cleaning; its validation set contains 1000 episodes. All episodes can be navigated without traversing between floors.

\underline{\textit{HM3D\_OVON}} \cite{yokoyama2024hm3d} broadens the scope and semantic range of prior Object Goal Navigation benchmarks. It covers 379 distinct categories and supports training and evaluation using goal sets defined via free-form language. Its validation set contains 3000 episodes.

\underline{\textit{MP3D:}} In the Habitat Challenge 2021 \cite{habitatChallenge2021}\cite{batra2020objectnav}, an Object Goal Navigation dataset was constructed based on MP3D scenes \cite{chang2017matterport3d}. It includes 21 annotated categories, and its validation set contains 2195 episodes. 

\textbf{Metrics.}
Across all datasets, we use Success Rate (SR) and Success weighted by Path Length (SPL) \cite{anderson2018evaluation} as evaluation metrics.
SR represents the percentage of successful episodes completions out of the total number of episodes. SPL quantifies the agent's trajectory efficiency by calculating the inverse ratio of the actual path length to the optimal path length weighted by the success rate.

\subsection{Case Study}
\begin{figure}[t]
  \centering
  \includegraphics[width=0.6\linewidth]{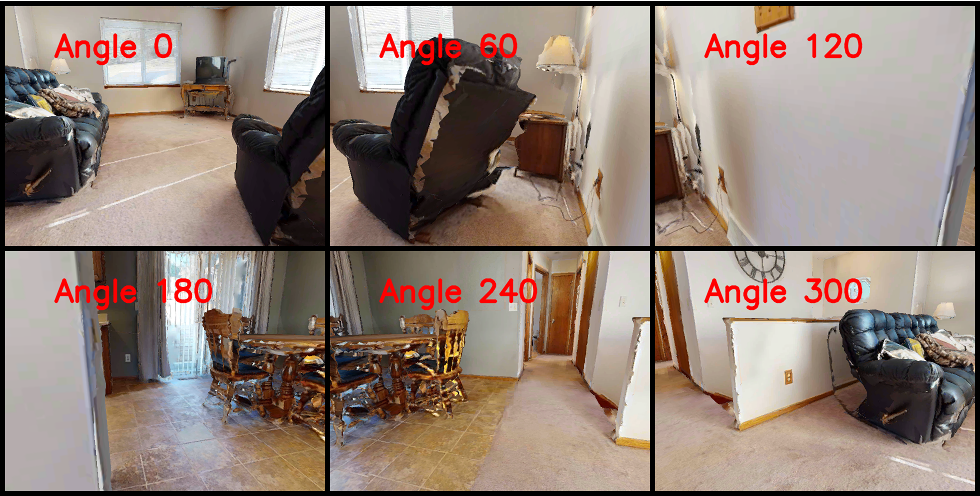}
  \caption{Panorama.}
  \label{fig:panorama}
\end{figure}

\textbf{Interpretability.}
The panorama retrieved by the Analysis Agent is shown in Fig. \ref{fig:panorama}, and its output is presented in the text box.
In the analysis of the 240° view, the model explicitly responds to the injected common-sense knowledge (“Please note that your body width is approximately 0.4 meters...”), indicating that common-sense injection encourages the model to provide clearer evidence along its reasoning pathway, thereby improving the transparency and interpretability of the decision-making process.

In the analysis of the 300° view, the model jointly considers image information from the local, global, and historical perspectives, demonstrating that these three sources play distinct yet complementary roles during the analysis. This suggests that multi-dimensional information fusion helps the model produce more reliable and higher-quality outputs in a more interpretable manner.

\begin{tcolorbox}[
  enhanced,
  breakable,
  colback=white,
  colframe=white,
  borderline west={2pt}{0pt}{black!30},
  left=6pt,right=6pt,top=4pt,bottom=4pt
]
\ttfamily\small
'240': \{'score': 6, 'reason': '...Since the direction aligns with an unexplored path and there’s no obstacle blocking forward movement \textbf{(width >0.4m)}, it’s moderately promising...'\}, \\
'300': \{'score': 3, 'reason': '\textbf{The view at 300 degrees} shows a partial wall and a large leather sofa, with no sign of a bed. \textbf{The panoramic view }confirms this is part of a living area, and \textbf{the top-down map} indicates that while the immediate vicinity (gray) has been explored...'\}
\end{tcolorbox}

\begin{figure}[t]
  \centering
  \includegraphics[width=0.6\linewidth]{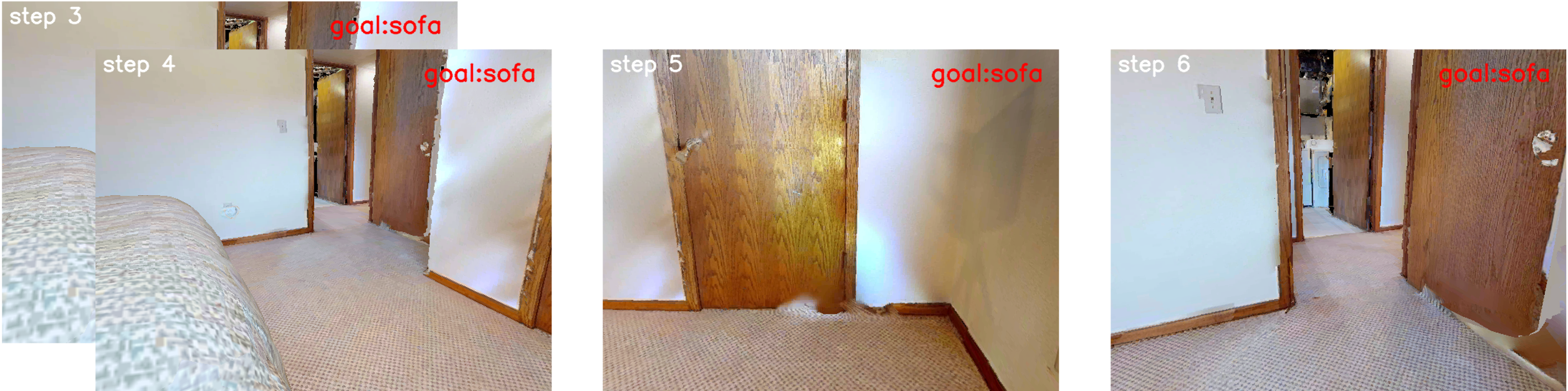}
  \caption{Jump over obstacles.}
  \label{fig:skip_obstacles}
\end{figure}

\textbf{Two-step Review.}
The Figure \ref{fig:skip_obstacles} illustrates a specific example of the two-step retrospective check. This check first monitors in real time whether the state at time t matches what was expected at time t–1. If the observed images from the two steps are essentially the same, this indicates that the outcome is not as expected, which triggers the obstacle-bypass mode. In this case, the previously selected best direction must be excluded, and the next-best direction is chosen instead. The system then continues moving until the obstacle is bypassed.

\begin{figure}[t]
  \centering
  \includegraphics[width=0.6\linewidth]{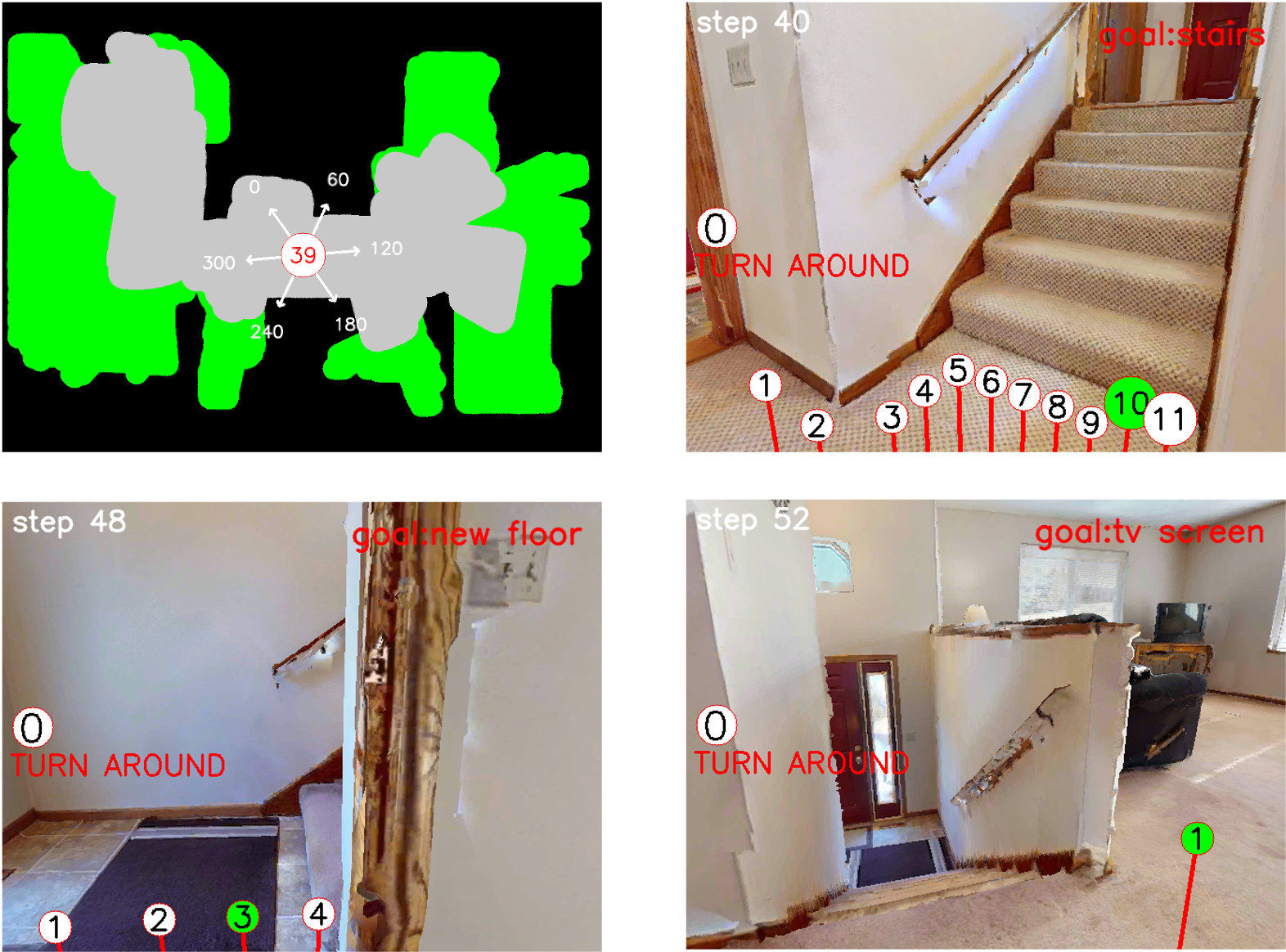}
  \caption{Jump out of floor.}
  \label{fig:skip_floor}
\end{figure}

\textbf{Multi-step Review.}
The Figure \ref{fig:skip_floor} illustrates a specific example of a multi-step retrospective check. When the agent finishes exploring the current floor but still fails to find the target, the “floor-exit” mode is triggered. It first searches for a staircase; once found, it climbs the stairs to reach a new floor. After arriving, it continues searching for the original task target.

\end{document}